\documentclass{article}
\usepackage{lmodern}
\usepackage{inconsolata}
\makeatletter
\def\@toptitlebar{\hrule height 4pt \vskip 0.25in \vskip -\parskip}
\def\@bottomtitlebar{\vskip 0.29in \vskip -\parskip \hrule height 1pt \vskip 0.09in}
\makeatother
\usepackage{graphicx}
\usepackage[authoryear]{natbib} 
\usepackage{booktabs}
\usepackage{url}
\usepackage[dvipsnames]{xcolor}
\usepackage[most]{tcolorbox}
\usepackage{authblk} 
\usepackage{placeins}
\usepackage{newunicodechar}
\usepackage{float} 
\newunicodechar{−}{\textminus}
\usepackage{inconsolata}

\usepackage[verbose=true,letterpaper]{geometry}
\AtBeginDocument{
  \newgeometry{
    textheight=9in,
    textwidth=6.5in,
    top=1in,
    headheight=14pt,
    headsep=25pt,
    footskip=30pt
  }
}
\usepackage{fancyhdr}
\fancyheadoffset{0pt}
\def\keywordname{{\bfseries \emph Keywords}}%
\def\keywords#1{\par\addvspace\medskipamount{\rightskip=0pt plus1cm
\def\and{\ifhmode\unskip\nobreak\fi\ $\cdot$
}\noindent\keywordname\enspace\ignorespaces#1\par}}
\makeatletter
\renewcommand{\normalsize}{%
  \@setfontsize\normalsize\@xpt\@xipt
  \abovedisplayskip      7\p@ \@plus 2\p@ \@minus 5\p@
  \abovedisplayshortskip \z@ \@plus 3\p@
  \belowdisplayskip      \abovedisplayskip
  \belowdisplayshortskip 4\p@ \@plus 3\p@ \@minus 3\p@
}
\normalsize
\renewcommand{\small}{%
  \@setfontsize\small\@ixpt\@xpt
  \abovedisplayskip      6\p@ \@plus 1.5\p@ \@minus 4\p@
  \abovedisplayshortskip \z@  \@plus 2\p@
  \belowdisplayskip      \abovedisplayskip
  \belowdisplayshortskip 3\p@ \@plus 2\p@    \@minus 2\p@
}
\renewcommand{\footnotesize}{\@setfontsize\footnotesize\@ixpt\@xpt}
\renewcommand{\scriptsize}{\@setfontsize\scriptsize\@viipt\@viiipt}
\renewcommand{\tiny}{\@setfontsize\tiny\@vipt\@viipt}
\renewcommand{\large}{\@setfontsize\large\@xiipt{14}}
\renewcommand{\Large}{\@setfontsize\Large\@xivpt{16}}
\renewcommand{\LARGE}{\@setfontsize\LARGE\@xviipt{20}}
\renewcommand{\huge}{\@setfontsize\huge\@xxpt{23}}
\renewcommand{\Huge}{\@setfontsize\Huge\@xxvpt{28}}
\providecommand{\section}{}
\renewcommand{\section}{%
  \@startsection{section}{1}{\z@}%
                {-2.0ex \@plus -0.5ex \@minus -0.2ex}%
                { 1.5ex \@plus  0.3ex \@minus  0.2ex}%
                {\large\bf\raggedright}%
}
\providecommand{\subsection}{}
\renewcommand{\subsection}{%
  \@startsection{subsection}{2}{\z@}%
                {-1.8ex \@plus -0.5ex \@minus -0.2ex}%
                { 0.8ex \@plus  0.2ex}%
                {\normalsize\bf\raggedright}%
}
\providecommand{\subsubsection}{}
\renewcommand{\subsubsection}{%
  \@startsection{subsubsection}{3}{\z@}%
                {-1.5ex \@plus -0.5ex \@minus -0.2ex}%
                { 0.5ex \@plus  0.2ex}%
                {\normalsize\bf\raggedright}%
}
\providecommand{\paragraph}{}
\renewcommand{\paragraph}{%
  \@startsection{paragraph}{4}{\z@}%
                {1.5ex \@plus 0.5ex \@minus 0.2ex}%
                {-1em}%
                {\normalsize\bf}%
}
\providecommand{\subparagraph}{}
\renewcommand{\subparagraph}{%
  \@startsection{subparagraph}{5}{\z@}%
                {1.5ex \@plus 0.5ex \@minus 0.2ex}%
                {-1em}%
                {\normalsize\bf}%
}

\newlength{\@abovecaptionskip}
\newlength{\@belowcaptionskip}
\renewenvironment{table}
  {\setlength{\abovecaptionskip}{\@belowcaptionskip}%
   \setlength{\belowcaptionskip}{\@abovecaptionskip}%
   \@float{table}}
  {\end@float}
\renewcommand{\footnoterule}{\kern-3\p@ \hrule width 12pc \kern 2.6\p@}
\def\@listi  {\leftmargin\leftmargini}
\def\@listii {\leftmargin\leftmarginii
                \labelwidth\leftmarginii
                \advance\labelwidth-\labelsep
                \topsep  2\p@ \@plus 1\p@    \@minus 0.5\p@
                \parsep  1\p@ \@plus 0.5\p@ \@minus 0.5\p@
                \itemsep \parsep}
\def\@listiii{\leftmargin\leftmarginiii
                \labelwidth\leftmarginiii
                \advance\labelwidth-\labelsep
                \topsep    1\p@ \@plus 0.5\p@ \@minus 0.5\p@
                \parsep    \z@
                \partopsep 0.5\p@ \@plus 0\p@ \@minus 0.5\p@
                \itemsep \topsep}
\def\@listiv {\leftmargin\leftmarginiv
                \labelwidth\leftmarginiv
                \advance\labelwidth-\labelsep}
\def\@listv  {\leftmargin\leftmarginv
                \labelwidth\leftmarginv
                \advance\labelwidth-\labelsep}
\def\@listvi {\leftmargin\leftmarginvi
                \labelwidth\leftmarginvi
                \advance\labelwidth-\labelsep}

\renewcommand{\maketitle}{%
  \par
  \begingroup
    \renewcommand{\thefootnote}{\fnsymbol{footnote}}
    \def\@makefnmark{\hbox to \z@{$^{\@thefnmark}$\hss}}
    \long\def\@makefntext##1{%
      \parindent 1em\noindent
      \hbox to 1.8em{\hss $\m@th ^{\@thefnmark}$}##1
    }
    \thispagestyle{empty}
    \@toptitlebar
    \centering
    {\LARGE\scshape \@title\par}
    \@bottomtitlebar
    \vspace*{.2in} 
    \begingroup\centering
      \def\And{\end{tabular}\hfil\linebreak\hfil\begin{tabular}[t]{c}}\def\AND{\end{tabular}\hfil\linebreak\hfil\begin{tabular}[t]{c}}%
      \begin{tabular}[t]{c}\bfseries\@author\end{tabular}\par%
    \endgroup\par
    \vskip 0.4in \@minus 0.1in \center{\today} \vskip 0.2in
  \endgroup
  \setcounter{footnote}{0}
  \let\maketitle\relax
  \let\@maketitle\relax
  \let\thanks\relax
  \let\@thanks\@empty
  \let\author\relax
  \let\@author\relax
  \let\title\relax
  \let\@title\relax
  \let\date\relax
  \let\@date\relax
}

\makeatother

\begin{document}

% --- Title, Author, Abstract ---
\title{Continual Learning, Not Training: Online Adaptation for Agents}

% --- AUTHOR ---
\author[1]{Aman Jaglan}
\author[1]{Jarrod Barnes}
\affil[1]{Arc Intelligence \\ \texttt{\{aman, jarrod\}@arc.computer}}
\affil[ ]{\small Corresponding author: \texttt{jarrod@arc.computer}}
% --- END AUTHOR BLOCK ---

\maketitle

% --- ABSTRACT ---
\begin{abstract}
Continual Learning (CL) methods have traditionally focused on mitigating catastrophic forgetting through gradient-based retraining, an approach ill-suited for deployed agents that must adapt in real time. We introduce our Adaptive Teaching and Learning System (ATLAS), a dual-agent architecture that decouples reasoning (Teacher) from execution (Student) and incorporates a persistent learning memory that stores distilled guidance from experience. This informs the orchestration layer, enabling the system to dynamically adjust its operational strategies, such as supervision level or initial plan selection, at inference time. In doing so, ATLAS achieves gradient-free continual learning, shifting the locus of adaptation from model parameters to system-level orchestration. We formulate this as a system-centric paradigm for continual learning, where the objective is adaptive efficiency: maximizing task success while minimizing computational cost through inference-time orchestration rather than parameter updates \citep{kirkpatrick2017overcoming, rolnick2019experience}. Evaluated on Microsoft’s ExCyTIn-Bench (Incident \#5 subset) \citep{wu2025excytin}, an open-source benchmark simulating complex cyber-threat investigation, ATLAS achieves 54.1\% success with GPT-5-mini as its Student, outperforming the larger GPT-5 (High) by 13\% while reducing cost by 86\%. Inference-time continual learning positions this approach on the Pareto frontier: superior accuracy at lower computational cost, achieved through gradient-free adaptation during deployment. The system demonstrates progressive efficiency gains across the 98-task trajectory, reducing token consumption from 100,810 (tasks 1–25) to 67,002 (tasks 61–98) while maintaining mid-50\% success, confirming that ATLAS learns to solve incidents more economically without sacrificing accuracy. Cross-incident validation on Incident \#55 demonstrates generalization: frozen pamphlets from Incident \#5 improve accuracy from 28\% to 41\% (+46\%) with zero retraining, while shifting output composition from verbose exploration to structured reasoning (−52\% non-reasoning tokens, +2,135 reasoning tokens per question). Together, these findings establish gradient-free continual learning as a viable path toward adaptive, deployable AI systems and provide causally annotated traces valuable for training explicit world models \citep{yu2023explainable}.
Code and reproducibility assets are available at \url{https://github.com/Arc-Computer/atlas-sdk}; the accompanying dataset will be released to enable replication and further study.
\end{abstract}


% --- ACKNOWLEDGMENTS ---
\section*{Acknowledgments}
The authors would like to thank \textbf{Michelangelo Naim} for their valuable contributions, discussions, and feedback throughout the development of this work. 
% --- END ACKNOWLEDGMENTS ---

% --- SECTION 1 ---
\section{Introduction}
Deployed language model agents operate in dynamic environments requiring continuous adaptation, yet their core knowledge remains static after pretraining \citep{pham2021dualnet}. This creates a tension, how can systems adapt in real-time when the dominant learning paradigms rely on offline training cycles? While Continual Learning (CL) addresses knowledge updates, existing approaches focus overwhelmingly on mitigating catastrophic forgetting through gradient-based weight updates conducted offline \citep{kirkpatrick2017overcoming, rolnick2019experience}, an inherently model-centric paradigm ill-suited for deployment constraints.

In complex adaptive systems, the environment perpetually evolves, by the time a model completes offline training on one configuration, the live system may have already shifted. Backpropagation, even in efficient forms like parameter-efficient fine-tuning (PEFT, e.g., LoRA; \citealt{hu2021lora}) or sparse-update methods, necessitates dedicated training loops, specialized hardware, data accumulation, and introduces retraining delays. These approaches cannot provide the inference-time adaptation needed for agents operating in real-time, often on resource-constrained hardware or without access to training infrastructure \citep{guo2021diff, dettmers2023qlora, liu2024dora, frantar2023sparsegpt}.

To address this challenge, we propose a system-centric approach to continual learning designed for inference-time deployment. Rather than focusing on updating weights without forgetting, we reframe the goal as achieving efficient performance: measurable improvements in task success and reductions in computational cost e.g., tokens consumed, as the system gains experience during live operation. This requires shifting the locus of adaptation from model weights to the system's orchestration layer.

We introduce ATLAS, a dual-agent architecture operationalizing this paradigm shift from model- to system-centric continual learning. ATLAS achieves gradient-free adaptation at inference-time through memory-guided orchestration, using aggregated learning history and rewards derived from Teacher-Student interactions to dynamically adjust its operational strategy. This mechanism requires no gradients, no model retraining, and no specialized hardware, making adaptation immediate, cost-efficient, transparent, and deployable by practitioners on standard inference hardware.

On Microsoft’s ExCyTIn-Bench (Incident \#5), ATLAS lifts success of GPT-5-mini from 33.7\% to 54.1\%, trims the Student’s tokens from 141,660 to 78,118 (−45\%), and surpasses the larger GPT-5 (High) by 13\%. This system-centric process serves as a novel data engine, demonstrating how its adaptive mechanism naturally generates the causally-annotated traces needed for training explicit world models \citep{yu2023explainable, ha2018world}.

% --- SECTION 2 ---
\section{Related Work}
The canon of current literature relating to CL falls under four main categories: (1) training-based approaches that suffer from catastrophic forgetting and require computationally expensive gradient updates \citep{kirkpatrick2017overcoming, rolnick2019experience, pham2021dualnet, hu2021lora, guo2021diff, dettmers2023qlora, liu2024dora, frantar2023sparsegpt}, (2) prompt optimization techniques that produce static instructions for deployment \citep{lester2021prompt, khattab2023dspy, agrawal2025gepa}, (3) retrieval-augmented systems that perform lookup rather than skill synthesis \citep{lewis2020rag, guu2020realm, borgeaud2022retro, asai2024self, lin2024radit}, and (4) agent memory mechanisms that passively store experiences without extracting generalizable knowledge \citep{shinn2023reflexion, zhou2024lats, wang2023voyager, packer2023memgpt}. Importantly none of these methods enable closed-loop, gradient‑free skill refinement during deployment. Building on existing work in continual learning, we argue that current approaches fail to support real-time adaptation, as they remain anchored to gradient-based retraining. We instead propose a system-centric framework that enables gradient-free, inference-time adaptation by decoupling reasoning from execution.

\subsection{Training-based}
The current dominant method in CL is training-based, with a focus on mitigating catastrophic forgetting via updating weights. Methods such as LoRA \citep{hu2021lora}, QLoRA \citep{dettmers2023qlora}, and DoRA \citep{liu2024dora} , sparse-update mechanisms, and experience replay techniques reduce computational costs but remain fundamentally constrained by their reliance on gradient-based optimization \citep{hu2021lora, guo2021diff, frantar2023sparsegpt, dettmers2023qlora, liu2024dora}. Even recent proposals for "fast-slow" dual-speed learning systems that implement rapid parameter updates still depend on gradient computation and cannot achieve immediate behavioral modification during task execution \citep{pham2021dualnet}.

These learning methods face a tension, while aggressive learning rates precipitate the instability of catastrophic forgetting, overly cautious updates restrict the agent's ability to achieve sufficient generalization and adaptation. Our system-centric approach with ATLAS, offers a departure from this method by treating model parameters as static and moving adaptation from the training loop into the inference-time orchestration of agent interactions, enabling immediate learning without gradient computation or forgetting.

\subsection{Prompt Optimization}
Representative systems include Prompt Tuning \citep{lester2021prompt}, DSPy compilation with self‑improving pipelines \citep{khattab2023dspy}, and recent GEPA evolutionary methods that outperform RL‑style optimizers in sample efficiency \citep{agrawal2025gepa}, while other methods leverage gradient-based prompt tuning or reinforcement learning over discrete prompt spaces. These techniques assess what is the optimal prompt for a task but are static and do not evolve in environmental conditions post-deployment.

ATLAS allows agents to continuously adapt their execution strategy based on a history of task-specific successes and failures. This stateful learning, which emerges dynamically during inference, enables highly context-specific specialization that addresses novel failure modes static prompts cannot handle.

\subsection{Retrieval Systems}
Retrieval mechanisms such as Retrieval-Augmented Generation (RAG) augment models by retrieving relevant documents or examples from external memory to provide the model with greater context \citep{lewis2020rag, guu2020realm, borgeaud2022retro}. More adaptive retrieval training such as Self‑RAG and RA‑DIT improves when/what to retrieve \citep{asai2024self, lin2024radit}.

ATLAS stores reward trajectories, which include past Teacher feedback, execution outcomes, and the corrective strategies employed. These trajectories are then utilized to train a meta-level control policy, enabling dynamic adjustments, such as varying the level of Teacher supervision based on the Student agent's performance patterns. This distinction shifts learning from the content level, knowledge augmentation to the strategic level, behavioral policy refinement, enabling skill acquisition rather than expanding content.

\subsection{Memory Mechanisms}
Episodic memory systems record interaction histories to improve future performance. Approaches such as Reflexion \citep{shinn2023reflexion}, LATS \citep{zhou2024lats}, Voyager \citep{wang2023voyager}, and MemGPT \citep{packer2023memgpt} maintain logs of past actions, outcomes, and textual reflections that agents can reference during subsequent tasks. This line of research implements memory as a passive, episodic log, a chronological record of observations and interactions. These systems lack a mechanism for active compression and generalization, memory grows with experience but past failures are not processed into generalizable knowledge that can steer future actions and alter execution policy. These systems also suffer from unbounded memory growth and lack principled consolidation and accumulate episodic traces without synthesizing higher-order behavioral rules.

In contrast, ATLAS implements memory as an active learning substrate, not a passive log. The Teacher's corrective feedback is explicitly structured and annotated, allowing the orchestrator to use this refined history to directly steer the Student's policy. This transformation enables progressive skill refinement and true procedural learning at inference time, all without modifying the model's underlying parameters.

% --- SECTION 3 ---
\section{Methodology}
To operationalize our system-centric paradigm for continual learning, we designed the ATLAS architecture specifically for gradient-free, inference-time adaptation. Its core components enable dynamic behavioral adjustments based on accumulated experience, directly addressing the goal of efficient performance without modifying underlying model weights. Our methodology centers on a dual-agent architecture engineered to decouple reasoning (Teacher) from execution (Student).

\begin{figure}[htbp]
  \centering
 
  \includegraphics[width=\textwidth]{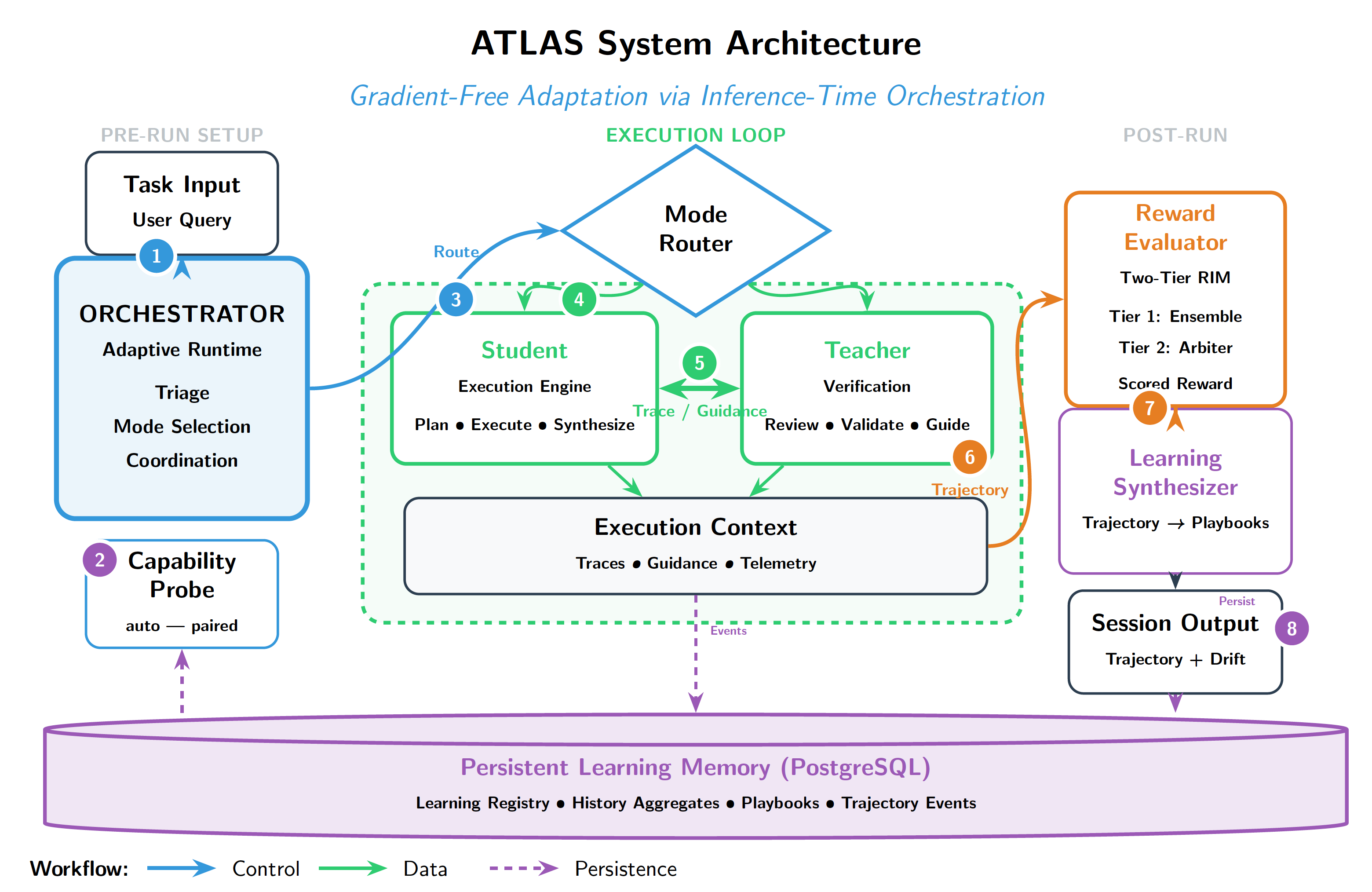}
  \caption{ATLAS System Architecture for gradient-free adaptation. An Orchestrator manages Teacher-Student interactions during the execution loop. Post-run, learning is evaluated, synthesized, and stored in Persistent Learning Memory (PLM) to guide future inference-time decisions.}
  \label{fig:system_design}
\end{figure}

\textbf{System Architecture:} The core of our system is an adaptive runtime orchestrating the interaction between a primary \textbf{Student} agent and a typically more capable \textbf{Teacher} agent. For each task, this interaction unfolds sequentially:
\begin{enumerate}
    \item \textbf{Task Execution:} The Student attempts the task, generating a trajectory of states, actions (e.g., tool calls, queries), and observations.
    \item \textbf{Guidance and Verification:} The Teacher observes the Student's trajectory. Based on task outcome and efficiency, it provides corrective, principle-level guidance.
    \item \textbf{Learning Persistence:} The runtime records the complete execution trace, Teacher guidance, and associated scores (derived from a two-tier reward system providing auditable rationales ) in a \textbf{Persistent Learning Memory (PLM)}, indexed by task context. A lightweight process distills actionable guidance from these records.
\end{enumerate}
This cycle enables Adaptive Learning at Inference Time. On subsequent similar tasks (identified by task context), the runtime retrieves the relevant aggregated learning history (distilled guidance and past performance) from the PLM. This retrieved information directly informs the orchestration layer, which dynamically adjusts the system's operational strategy, for example, selecting the appropriate Teacher supervision level (e.g., fully autonomous for high-confidence tasks, step-by-step guidance for novel ones) or seeding the Student's plan.

This entire adaptation process occurs purely at inference time, leaving model weights unchanged. Learning is mediated solely through the structured history stored in the PLM and retrieved by the orchestrator, enabling the system to develop "fast paths" for familiar scenarios while invoking escalated supervision for novel or challenging ones, thus transforming raw interaction traces into actionable, reusable learning.

\subsection{Inference-Time Learning}
Adaptation arises from a mechanistic loop. When the Student fails or demonstrates inefficiency, the Teacher provides concise, principle-level guidance, such as verifying the source IP and authentication path before conducting privilege analysis in a security triage task. The reward subsystem then evaluates both the Student’s trajectory and the Teacher’s intervention along predefined axes including factuality, instruction-following, efficiency, and safety, attaching structured rationales to each score. The learning engine subsequently compiles two artifacts: a Teacher Pamphlet, which captures principles, failure modes, diagnostics, and stop conditions; and a Student Pamphlet, which encodes the corresponding action schema, tool plan, guards, and success checks.

\subsection{Reward System}
To transform raw interactions into high-fidelity supervision, ATLAS employs a two-tier, ensemble-of-judges rewarder. Multiple fast judges independently score trajectories and guidance, each required to state the evaluation principles before assigning scores; when variance or self-reported uncertainty exceeds thresholds, a stronger arbiter consolidates the rationales and issues the final judgment \citep{jung2025trust} This routing preserves low cost on routine cases while allocating deliberation to ambiguous ones, and the resulting principle-grounded rationales form an audit trail for prompt and threshold tuning. Publicly reported results indicate this system attains high accuracy on RewardBench 2 \citep{malik2025rewardbench} via an ensemble-then-arbiter design, consistent with our use as an auditable reward signal within ATLAS.

\subsection{World-Model Data Engine}
A direct consequence of the ATLAS learning loop is the generation of structured and causally annotated data suitable for world-model training \citep{yu2023explainable}. Each execution trace contains three components: (1) state: task context, environment state, and intermediate observations; (2) action: the Student’s tool/API/SQL calls, plans, and decision points; and (3) outcome: observations, success and failure signals, artifacts, latencies, and retries. These traces also include the Teacher’s diagnostic guidance explaining why particular actions failed or succeeded, along with meta-signals such as supervision-lane decisions, confidence estimates, disagreement, and escalation events. Unlike raw or success-only logs, these traces provide explicit causal explanations for action outcomes and cover both optimal and correction-rich trajectories. We hypothesize that world models trained on ATLAS traces will exhibit improved predictive fidelity and sample efficiency relative to models trained on conventional trajectory data.

% --- SECTION 4 ---
\section{Experimental Setup}
\subsection{Benchmark}
The experiments are conducted on ExCyTIn-Bench \citep{wu2025excytin}, a cyber-threat investigation benchmark designed for stateful reasoning. As CL shifts evaluation away from static test sets, ExCyTIn-Bench offers a more process-aware assessment by scoring trajectories within a simulated incident environment. We focus on Incident \#5, which provides a consistent scenario and scoring protocol.

\subsection{System Configuration and Baselines}
\textbf{System Configuration}

Our experimental setup consists of two phases: seeding and evaluation.
During the seeding phase, we employ a paired configuration where GPT-5 serves as the Teacher model and GPT-5-mini as the Student model. Working on ExCyTIn-Bench Incident \#5 (98 queries), the Teacher observes Student trajectories and provides targeted guidance. These interactions are then distilled into Learning Pamphlets and stored in the Persistent Learning Memory (PLM).

In the evaluation phase, we assess cross-task transfer by retrieving relevant pamphlets via semantic similarity to initialize subsequent tasks. This tests whether guidance learned from earlier queries can improve performance on later, related queries within the same incident domain, without any model weight updates, maintaining the inference-time learning paradigm.

\textbf{Baseline}

We establish two comparison points to isolate the contribution of our approach:
\begin{enumerate}
    \item Internal baseline (Student-only): GPT-5-mini operating without pamphlet or Teacher guidance. This isolates the impact of our inference-time learning mechanism by measuring raw Student performance.
    \item External baseline (Benchmark reference): The reported GPT-5 (Reasoning = High) performance on Incident \#5 from ExCyTIn-Bench documentation, which achieved an average reward of 0.501. This provides a reference point from a stronger model using the benchmark's standard evaluation protocol.
\end{enumerate}
All experiments use identical reward configurations and evaluation protocols to ensure standardized comparison across baselines and our system.

\subsection{Metrics}
We measure performance across two metrics:
\begin{enumerate}
    \item \textbf{Task Success Rate:} binary success rate computed using the benchmark’s official criterion, reported both overall and on the benchmark’s flagged query subset, as defined by ExCyTIn-Bench. This provides a strict pass or fail assessment of task completion. Success is determined using ExCyTIn-Bench’s binary correctness evaluation, which applies a threshold of $\ge$0.4 on the ensemble reward score to account for judge uncertainty while maintaining strict answer verification.
    \item \textbf{Efficiency:} average tokens consumed per session, measuring the computational cost of achieving task objectives.
\end{enumerate}

% --- SECTION 5 ---
\section{Results}
We evaluate the paired Teacher–Student configuration using ExCyTIn-Bench's scoring protocol on Incident \#5 (n = 98 queries), with all models operating at inference time without weight updates.

\subsection{Efficiency Gains}
\textbf{Increasing token reduction}

\begin{figure}[htbp]
  \centering

  \includegraphics[width=\textwidth]{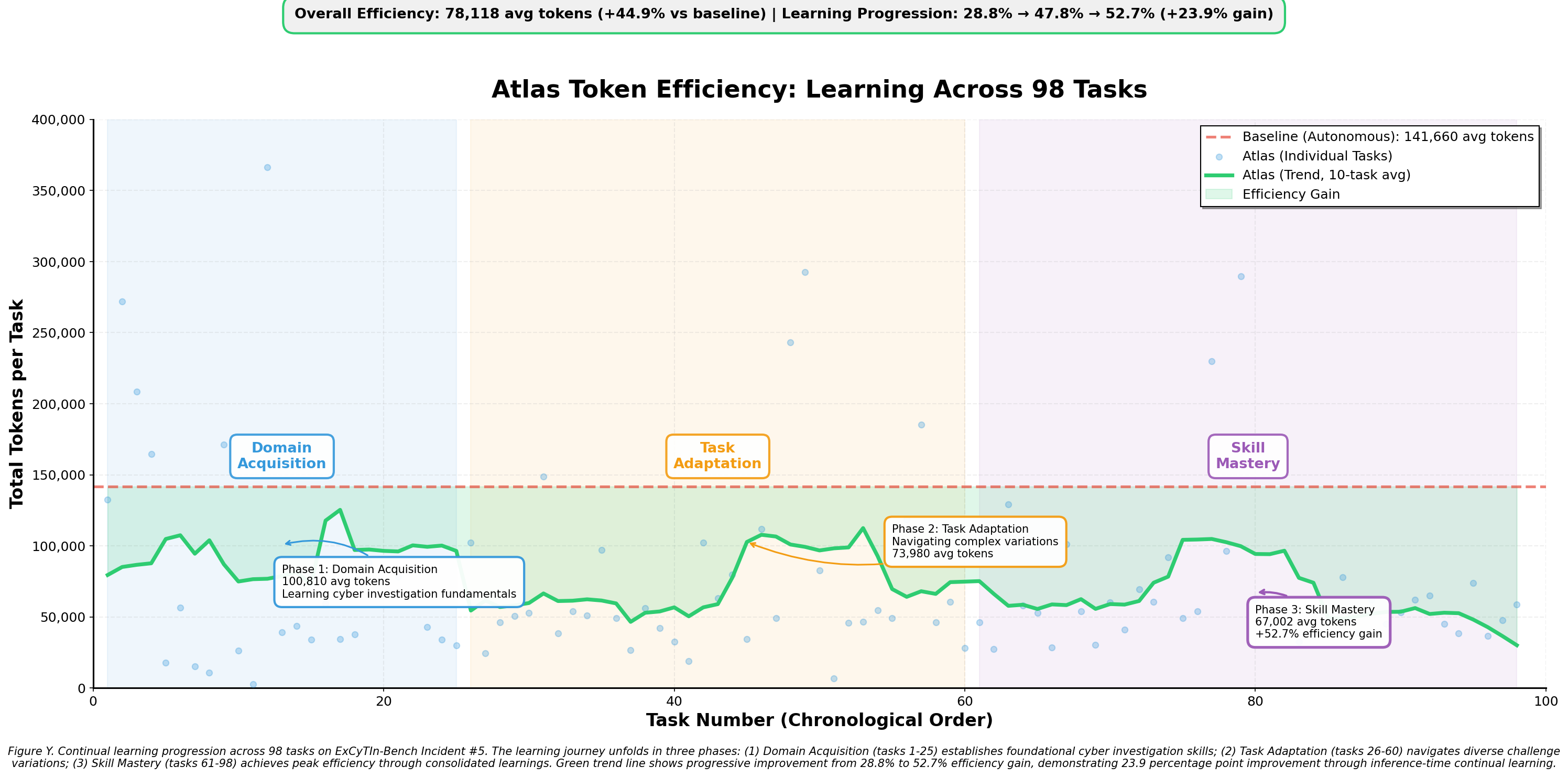}
  \caption{Continual learning progression across 98 tasks on ExCyTIn-Bench Incident \#5. The learning journey unfolds in three phases: (1) Domain Acquisition (tasks 1-25) establishes foundational cyber investigation skills; (2) Task Adaptation (tasks 26-60) navigates diverse challenge variations; (3) Skill Mastery (tasks 61-98) achieves peak efficiency through consolidated learnings. Green trend line shows progressive improvement from 32.8\% to 53.9\% efficiency gain, demonstrating 21.1 percentage point improvement through inference-time continual learning.}
  \label{fig:token_efficiency}
\end{figure}

ATLAS demonstrates systematic efficiency improvements as learned artifacts accumulate in the PLM and are retrieved in subsequent episodes (Figure \ref{fig:token_efficiency}). Overall, ATLAS averages 78,118 tokens per task, a 45\% reduction relative to the autonomous GPT-5-mini baseline (141,660 tokens averaged over the 42 logged runs out of 47).

Phase breakdown reveals consistent learning progression:
\begin{itemize}
    \item \textbf{Phase 1} (tasks 1–25): 100,810 tokens/task (−28.8\% vs. baseline)
    \item \textbf{Phase 2} (tasks 26–60): 73,980 tokens/task (−47.8\% vs. baseline)
    \item \textbf{Phase 3} (tasks 61–98): 67,002 tokens/task (−52.7\% vs. basline)
\end{itemize}

\textbf{Performance improvement}

Efficiency gains coincide with stable mid-50\% success: \textbf{52.0\% $\to$ 57.1\% $\to$ 52.6\%}, demonstrating that ATLAS maintains accuracy while spending fewer tokens. The green trend line in Figure \ref{fig:token_efficiency} reflects efficiency gains rising from 28.8\% in Phase 1 to 52.7\% in Phase 3, a +23.9 percentage point improvement. This demonstrates that reductions in token usage did not compromise accuracy, but instead facilitated more efficient and effective problem-solving with ATLAS.

\subsection{Benchmark Performance}
On Incident \#5, ATLAS achieves \textbf{54.1\% task success rate} (53/98 tasks) under the benchmark's scoring protocol. Compared to the GPT-5 (High) benchmark reference:
\begin{itemize}
    \item \textbf{+6.1 percentage point higher success rate} (54.1\% vs. 48.0\% for GPT-5 High)
    \item \textbf{\textasciitilde86\% lower dollar cost per question} ($\approx$\$0.024 vs. \$0.174 per question), using OpenAI’s published GPT-5-mini and GPT-5 pricing tiers applied to each model’s measured token usage.
\end{itemize}
These results demonstrate that ATLAS using a smaller model with inference-time learning, exceeds the performance of a larger baseline model at substantially lower computational cost \citep{wu2025excytin}.

% --- IMAGE ---
\begin{figure}[htbp]
  \centering
  \includegraphics[width=\textwidth]{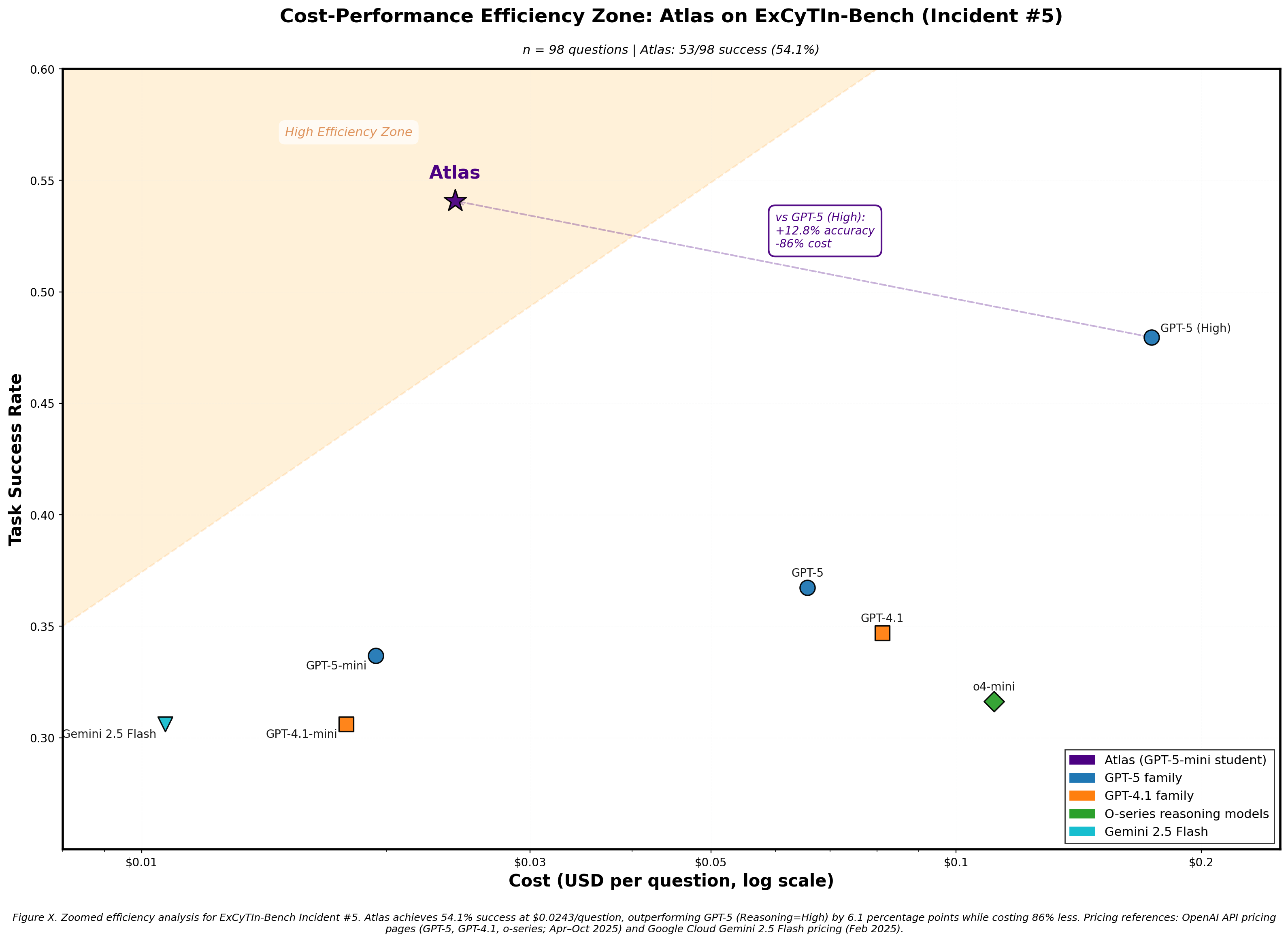}
  \label{fig:cost_performance}
  \caption{Performance and cost comparison on ExCyTIn-Bench (Incident \#5). ATLAS (blue) achieves higher success (54.1\%) at a lower cost (\textasciitilde\$0.024/task) than the stronger GPT-5 (High) baseline (48.0\%, \textasciitilde\$0.174/task).}
  \label{fig:pareto_zoom}
\end{figure}

% --- TABLE 1 ---
\begin{table}[H]
  \centering
  \caption{ExCyTIn-Bench (Incident \#5) Performance Summary}
  \label{tab:results}
  \renewcommand{\arraystretch}{1.5}
  \small
  \begin{tabular}{lrrl}
    \toprule
    \textbf{Configuration} & \textbf{Success (Overall)} & \textbf{Avg Tokens / Task} & \textbf{Notes} \\
    \midrule
    \textbf{ATLAS (Teacher \& Student)} & \textbf{54.1\%} & \textbf{78,118} & GPT-5-mini student guided by GPT-5 teacher (n=98) \\
    GPT-5 (Reasoning=High) & 48.0\% & 71,105 & Reproduced Incident \#5 run (n=98) \\
    GPT-5-mini (official baseline) & 33.7\% & 61,562 & Microsoft Incident \#5 release (n=98) \\
    GPT-5-mini (Student-only) & 40.4\% & 141,660$^{\dagger}$ & No pamphlets/teacher (n=47) \\
    \bottomrule
  \end{tabular}
  \parbox{\linewidth}{\scriptsize $^{\dagger}$Token averages computed over the 42/47 tasks with logged usage; success rate still counts all 47 tasks.}
\end{table}

% --- SECTION ---
\subsection{Cross-Incident Transfer (Incident 55)}
To test whether guidance distilled on Incident \#5 generalizes to new scenarios, we froze the learning memory and switched the runtime to auto mode (Student-only, no Teacher or Reward System). Running the same GPT-5-mini model on Incident \#55, the official baseline clears 28 of 100 questions. With stored pamphlets injected into the Student's context, but no new Teacher feedback or reward signals, the model answers 41 correctly, a +46\% improvement achieved purely through reused artifacts. This demonstrates that ATLAS learns transferable investigative strategies, not task-specific templates.

Output token analysis reveals efficient adaptation: completion tokens increase 50.3\% (from 2,085 to 3,134 per question), but the composition shifts dramatically. The baseline produces 2,085 non-reasoning output tokens per question, while ATLAS generates only 999 non-reasoning tokens (−52.1\%) and 2,135 reasoning tokens. Pamphlets guide the model toward deliberate reasoning rather than verbose exploration, reducing wasteful generation while increasing structured problem-solving. The output cost rises \$0.002 per question, yielding 13 additional correct answers, or \$0.016 per incremental success. While input prompt optimization remains necessary, the 28 $\to$ 41 accuracy gain provides our first empirical evidence that learning generalizes beyond the original incident domain.

% --- FIGURE ---
\begin{figure}[htbp]
  \centering
  \includegraphics[width=\textwidth]{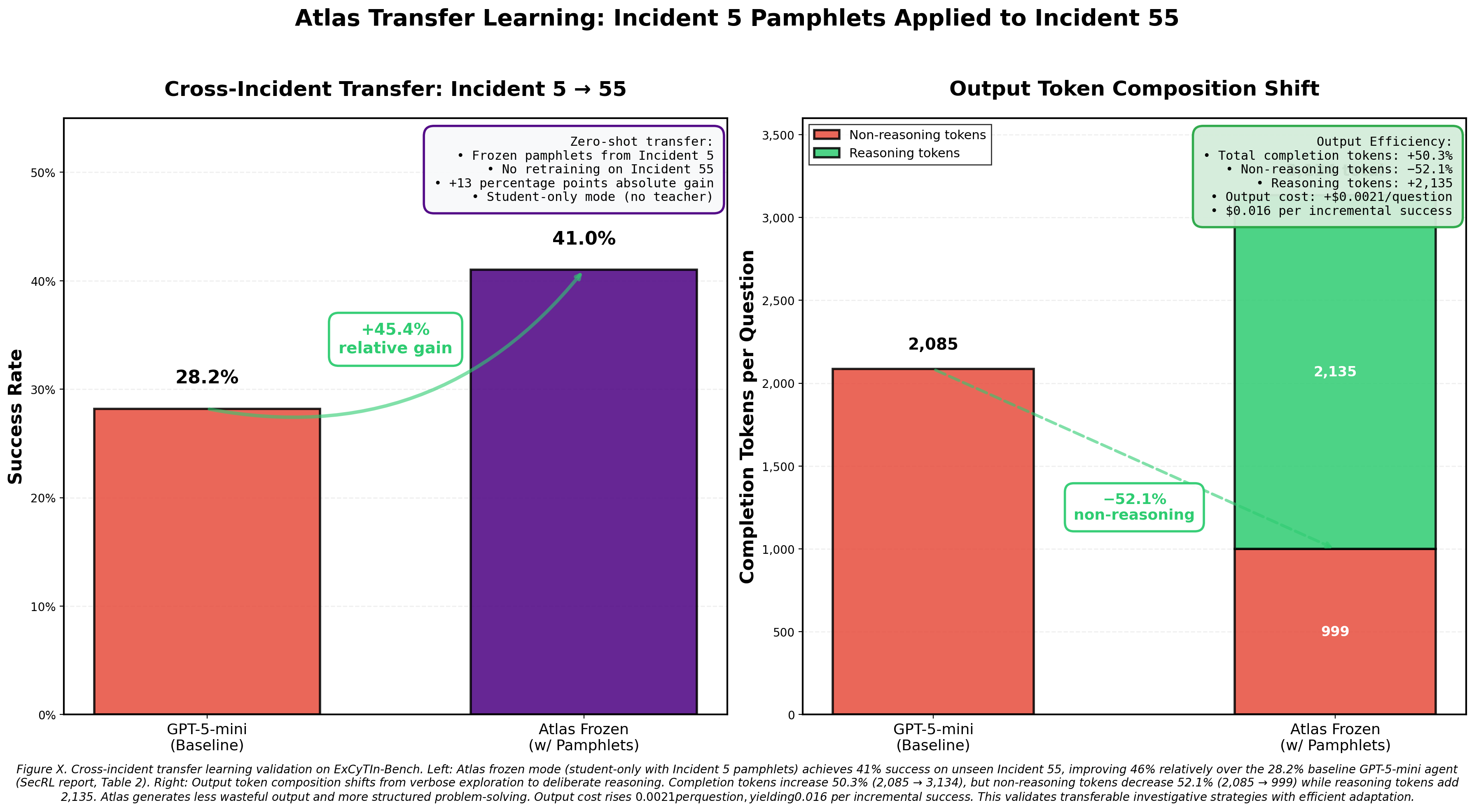}
  \caption{Cross-incident transfer performance from Incident \#5 to Incident \#55. Using learned pamphlets from Incident \#5, ATLAS improved success rate from 28\% to 41\% (+46\%) with no additional training or real-time guidance.}
  \label{fig:transfer_learning}
\end{figure}

% --- SECTION 6 ---
\section{Analysis}
The observed efficiency and accuracy gains in ATLAS arise from two mechanisms: adaptive teaching and distilled experience transfer (DET).

\textbf{Adaptive Teaching} provides dynamic, context-aware guidance during task execution. By observing the Student's trajectory in real-time, the Teacher agent identifies and flags low-yield exploratory paths early, suggests high-value strategic pivots (e.g., schema-guided SQL filtering), and provides principle-level corrective guidance when the Student stalls. This adaptive supervision prunes the search space, preventing wasted computation on unproductive actions. The reward subsystem further reinforces effective strategies by associating Teacher interventions with principle-grounded rationales, helping to shape more efficient decision-making patterns over time. This mechanism mirrors interactive teaching, where support is adjusted based on the learner's immediate needs \citep{liu2023geval, jung2025trust}.

\textbf{Distilled Experience Transfer (DET)} enables cross-task learning and accelerates adaptation by leveraging codified knowledge from past interactions. Actionable guidance and successful strategies, distilled by a lightweight process from Teacher interventions and high-reward trajectories, are stored as artifacts in the Persistent Learning Memory (PLM), indexed by task context. On subsequent, similar tasks, the orchestration layer surfaces this relevant distilled experience from the PLM. Applying this knowledge seeds the Student's initial plan or informs the runtime selection of the appropriate supervision level, allowing the system to bypass redundant exploration and rapidly apply proven tactics ("fast paths"). DET ensures that learnings from one episode are effectively transferred to future, related scenarios.

Together, \textbf{Adaptive Teaching} (real-time guidance) and \textbf{Distilled Experience Transfer} (leveraging past distilled lessons) create a synergistic effect. They enable the system to progressively shorten solution trajectories, reduce token consumption (Figure \ref{fig:token_efficiency}), and increase task success rates (Figures \ref{fig:cost_performance}, \ref{fig:pareto_zoom}).

Across the 98-task run, total tokens per task fall from 100,810 in the Domain Acquisition phase (tasks 1–25) to 67,002 in Skill Mastery (tasks 61–98) while success holds approximately 52–57\%, demonstrating that ATLAS learns to solve cybersecurity incidents more economically without trading away accuracy.

% --- SECTION 6.1 ---
\begin{tcolorbox}[
  colback=gray!5,
  colframe=black,
  fonttitle=\bfseries,
  title=Qualitative Example: Adaptation within a Single Incident
]
We trace the complete learning cycle through Incident \#5, session 71. The task required identifying the Security Identifier (SID) associated with suspicious remote activity on host \texttt{vnevado-win10r}.
\par
\textbf{Initial failure.} The Student’s first attempt produced an unverified answer and omitted the required structured reasoning trace. Critically, it failed to systematically inspect Windows security and incident telemetry tables, demonstrating a lack of principled investigation strategy.
\par
\textbf{Teacher intervention.} Observing this failure, the Teacher issued principle-level guidance:
\begin{itemize}
  \item Enumerate relevant telemetry sources before attempting attribution
  \item Prioritize tables: \texttt{DeviceProcessEvents}, \texttt{DeviceNetworkEvents}, \texttt{SecurityAlert}
  \item Join on host and trace identifiers; verify SID presence in returned records
\end{itemize}
The reward system evaluated this guidance with an auditable rationale and positive score. The learning engine then distilled the intervention into two complementary pamphlets:
\begin{itemize}
  \item \textbf{Teacher pamphlet:} High-level investigative principles + diagnostic patterns
  \item \textbf{Student pamphlet:} Concrete SQL action schemas + validation guards
\end{itemize}
\textbf{Successful re-execution.} When the task was re-attempted with retrieved pamphlets seeding the context, the Student executed a systematic approach:
\begin{enumerate}
  \item Issued \texttt{SHOW TABLES} and \texttt{DESCRIBE DeviceProcessEvents} to survey available schema
  \item Filtered by target host: \texttt{DeviceName='vnevado-win10r'} with relevant temporal bounds
  \item Constructed joins across \texttt{DeviceProcessEvents}, \texttt{DeviceNetworkEvents}, and \texttt{SecurityAlert}
  \item Extracted and verified the correct SID: \texttt{S-1-5-21-1840191660-8534830288-125585561-1522}
\end{enumerate}
The corrected trajectory satisfied the benchmark’s success criterion while consuming fewer tokens than the initial attempt. A parallel autonomous execution of the same “SID of the account involved in the suspicious remote activity” prompt exhausted 304,389 tokens without ever landing on the canonical SID, underscoring that the retrieved pamphlet injected a reusable investigative pattern (schema scan $\to$ host filter $\to$ joined evidence check) rather than a cached answer. This principle-constrained search, guided by stored artifacts, is representative of the aggregate improvements observed across all 98 tasks in Incident \#5.
\end{tcolorbox}

% --- SECTION 6.2 ---
\subsection{Cross-Task Transfer Patterns}
Trajectory analysis reveals systematic reuse of learned principles across heterogeneous tasks:
\begin{itemize}
    \item \textbf{Guidance reuse spans disparate task types.} 69 of 98 trajectories include retrieved guidance in \texttt{metadata.secrl\_applied\_guidance}, and 68 of those inject skills (schema hygiene, constraint reconciliation, format discipline) that are absent from the new prompt text, showing that pamphlets capture abstract procedures rather than task-specific templates.
    \item \textbf{Overlap with autonomous runs shows true transfer.} On the 42 prompts ATLAS succeeds 57.1\% of the time while consuming 83.6k tokens on average, versus 45.2\% success at 144.6k tokens for the teacher-free baseline. Notably, 33 of those paired runs leveraged stored guidance, so higher accuracy derives from retrieved principles applied to new executions, not from repeated answers.
    \item \textbf{Process investigations become progressively cheaper.} The first three process-centric questions (indices 1, 6, 11) consume 217.7k tokens on average, whereas the last three (indices 90, 95, 96) finish in 48.8k tokens. Each later trajectory begins with the schema/prompting checklist minted during early failures, demonstrating that the same pamphlet shortens very different process-forensics tasks over time.
\end{itemize}
Together these measurements show that ATLAS stores reusable investigative strategies in its PLM and redeploys them across heterogeneous prompts, which in turn explains the steady efficiency gains in Figure \ref{fig:token_efficiency}.

% --- SECTION 6.3 ---
\subsection{Reproducibility}
To enable independent verification and extension of our results, we have released a comprehensive supplementary dataset. This dataset includes complete session traces documenting timestamped action sequences across all 98 tasks, alongside full Teacher interventions that capture the rationales and guidance provided at each decision point. We provide reward annotations containing principle-grounded justifications paired with numerical scores, as well as all generated Learning Pamphlets with their associated metadata. Finally, we include detailed token accounting logs that align precisely with the benchmark’s official scoring configuration, permitting complete audit of our reported metrics and decision paths.

% --- SECTION 7 ---
\section{Future Research}
Our work on system-centric CL opens several research directions that can be explored, particularly given its unique advantages in accessibility, deployability, and practical real-world applicability. We highlight four key areas in the following subsections.

\subsection{Architectural Design Exploration}
We will continue to compare and study alternative system-based designs remains essential. Comparative studies examining multi-agent ensembles, hierarchical memory structures, and varied capability probing mechanisms would elucidate trade-offs between learning speed, computational overhead, and architectural complexity. A particularly compelling question is: Can corrective strategies learned by one system be transferred or hierarchically combined across different architectures?

\subsection{Knowledge Generalization}
The principles stored in ATLAS's persistent memory suggest opportunities for cross-model and cross-task generalization. Can Teacher-generated corrective feedback trained on one Student model accelerate adaptation when transferred to other agents? Developing principled methods for distilling, validating, and transferring learned strategies could dramatically enhance the portability and scalability of system-centric learning, potentially creating reusable "libraries" of adaptive principles.

\subsection{Adaptive Evaluation Methodologies}
As we build systems that learn continuously, static benchmarks become insufficient. Our future work will focus on creating dynamic benchmarks that adapt alongside the agent, potentially increasing difficulty or introducing novel scenarios based on agent performance, and developing robust metrics to measure adaptation beyond simple task success, such as resilience to distribution shifts or efficiency of knowledge acquisition as mitigating evaluation hacking remains a key challenge.

\subsection{Hybrid Online and Offline Learning}
Another promising research direction lies in integrating world models trained offline on ATLAS-generated execution traces back into the live system. These learned models could serve as predictive simulators for counterfactual planning, as structured knowledge sources augmenting Teacher reasoning, or as components enabling more sophisticated credit assignment. Such world models could serve as planning simulators, Teacher enhancement modules, or sources of structured causal priors. By closing this loop between online and offline learning, ATLAS could evolve into a hybrid continual learner, one that couples immediate, inference-time adaptation with deeper, model-based understanding \citep{yu2023explainable}.

% --- SECTION 8 ---
\section{Conclusion}
We challenge the dominant model-centric paradigm in Continual Learning (CL), arguing that it fails to meet the demands of dynamic, real-world deployment. In such environments, adaptation cannot depend on offline retraining. True adaptability must occur at inference time.

We proposed system-centric CL with the defined goal of achieving efficient performance, measurable improvements in task success together with reductions in computational cost at inference-time.

We introduced ATLAS, a dual-agent architecture that operationalizes this paradigm. By orchestrating Student and Teacher models around a persistent learning memory, ATLAS achieves gradient-free adaptation, bypassing the need for retraining, specialized hardware, or adaptation-specific datasets. This democratizes continual learning, making it accessible for deployment on standard inference infrastructure.

Our evaluation on ExCyTIn-Bench validates this approach. ATLAS improved task success from 33.7\% (benchmark GPT-5-mini) / 40.4\% (Student-only) to 54.1\% as experience accumulated, while reducing the Student’s average tokens from 141,660 to 78,118 (−45\%). The adapted system using GPT-5-mini as Student achieved 54.1\% task completion, surpassing the 48.0\% baseline of the larger GPT-5 (High) model while operating at the GPT-5-mini cost tier. These results demonstrate that system-centric learning matches or exceeds compute-intensive training methods under strict inference-time constraints.

Beyond performance, our work highlights the urgent need for adaptive evaluation methods capable of assessing dynamically learning systems, a critical gap as the field moves beyond static models. Furthermore, we demonstrated that the causally-annotated traces generated by ATLAS’s learning process provide a powerful data engine for training world models, bridging online adaptation and offline model building.

In summary, system-centric CL offers a accessible, scalable, and immediately deployable path toward AI systems that improve through use learning efficiently and incrementally within the environments they are deployed.

% --- BIBLIOGRAPHY ---

\end{document}